\pdfoutput=1

\documentclass[11pt]{article}

\usepackage[]{ACL2023}

\usepackage{times}
\usepackage{latexsym}

\usepackage{microtype}
\usepackage{setspace}

\usepackage{graphicx}
\usepackage{soul}
\usepackage{url}
\usepackage{hyperref}
\usepackage{amsmath}
\usepackage{amsthm}
\usepackage{booktabs}
\usepackage{algorithm}
\usepackage{algorithmic}
\usepackage{array}
\usepackage{color, colortbl}
\usepackage{multirow}
\usepackage{setspace}
\usepackage{autobreak}
\usepackage{amsfonts}
\usepackage{amssymb}
\usepackage{amsthm,amsmath}
\usepackage{mathrsfs}
\usepackage{multirow}
\usepackage{subfigure}
\usepackage{makecell}
\usepackage{pifont}
\usepackage{fontawesome}
\usepackage{bm}

\usepackage[T1]{fontenc}

\usepackage[utf8]{inputenc}

\usepackage{microtype}

\usepackage{inconsolata}

\title{On the Universal Adversarial Perturbations for Efficient Data-free Adversarial Detection}

\author{
    Songyang Gao$^{1}$, \ \ Shihan Dou$^{1}$, \ \  Qi Zhang$^{12}$\thanks{{ }{ }{ }Corresponding author.}, \ \ Xuanjing Huang$^{12}$, \ \   Jin Ma$^3$, \ \ Ying Shan$^3$ \\ 
    \normalsize{$^1$  School of Computer Science, Fudan University,\ Shanghai,\ China} \\
    \normalsize{$^2$  Shanghai Key Laboratory of Intelligent Information Processing,\ Shanghai,\ China} \\
    \normalsize{$^3$Tencent PCG} \\
    \normalsize{ \{gaosy21, shdou21\}@m.fudan.edu.cn}
}

\begin{document}
\maketitle

\begin{abstract}
Detecting adversarial samples that are carefully crafted to fool the model is a critical step to socially-secure applications. However, existing adversarial detection methods require access to sufficient training data, which brings noteworthy concerns regarding privacy leakage and generalizability. 
In this work, we validate that the adversarial sample generated by attack algorithms is strongly related to a specific vector in the high-dimensional inputs.
Such vectors, namely UAPs (Universal Adversarial Perturbations), can be calculated without original training data. Based on this discovery, we propose a data-agnostic adversarial detection framework, which induces different responses between normal and adversarial samples to UAPs.
Experimental results show that our method achieves competitive detection performance on various text classification tasks, and maintains an equivalent time consumption to normal inference.

\end{abstract}

\section{Introduction}

Despite remarkable performance on various NLP tasks, pre-trained language models (PrLMs), like BERT \cite{devlin2018bert}, are highly vulnerable to adversarial samples \cite{zhang2020adversarial, zeng2021openattack}. 
Through intentionally designed perturbations, attackers can modify the model predictions to a specified output while maintaining syntactic and grammatical consistency \cite{jin2020bert, li2020bert}. Such sensitivity and vulnerability induce persistent concerns about the security of NLP systems \cite{zhang2021crafting}. Compared to deploying robust new models, it would be more applicable to production scenarios by distinguishing adversarial examples from normal inputs and discarding them before the inference phase \cite{shafahi2019adversarial}. Such detection-discard strategy helps to reduce the effectiveness of adversarial samples and can be combined with existing defence methods \cite{mozes2021frequency}.

\begin{figure}[t]
\centerline{\includegraphics[width=0.45\textwidth]{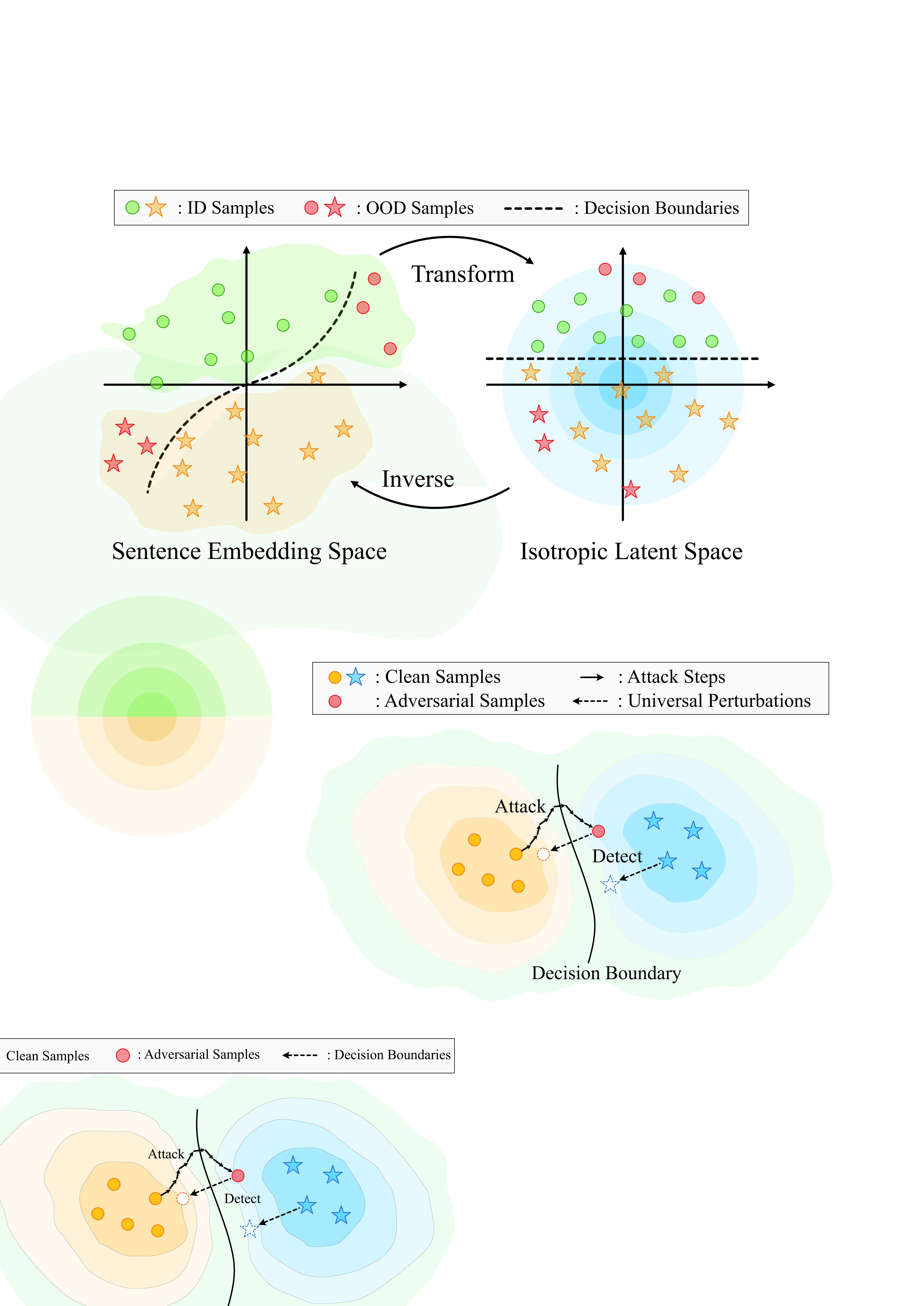}}
\caption{Illustration of our UAPAD framework. The solid and hollow markers represent samples before and after the universal perturbation. The adversarial samples are embedded closer to the decision boundary to maintain similarity with the original samples, resulting in differential resistance to universal adversarial perturbations (UAPs) with clean samples. We construct our detection framework based on this observation.}
\label{fig:1}
\vspace{-1.2em}
\end{figure}


However, existing adversarial detection methods depend heavily on the statistical characteristics of the training data manifolds, 
such as density estimation \cite{yoo2022detection} and local intrinsic dimensionality \cite{liu2022detecting}. Some other researches focus on identifying high-frequency words in the training data and replacing or masking them in the prediction phase to observe the change in logits score \cite{mozes2021frequency, mosca2022suspicious}.
We propose a summary of existing works in Table 1. All these detection methods assume that training data is available, which suffers from the following two problems:
(1) Some companies only provide model checkpoints without customer data due to privacy and security issues.
(2) Some datasets can be large so it is not practical or convenient to save and process them on different platforms.

\begin{table*}[th]
\resizebox{\textwidth}{!}{%
\centering
\begin{tabular}{llccc}
\hline
\textbf{Method} & \textbf{Summary} & \multicolumn{1}{c}{\textbf{\begin{tabular}[c]{@{}c@{}}Require \\ Clean Data\end{tabular}}} & \multicolumn{1}{c}{\textbf{\begin{tabular}[c]{@{}c@{}}Require \\ Adv. Data\end{tabular}}} & \multicolumn{1}{c}{\textbf{\begin{tabular}[c]{@{}c@{}}Require\\ Extra Model\end{tabular}}} \\ \hline
MLE \cite{lee2018simple} & Gaussian discriminant analysis &  \color[RGB]{200,85,80}\ding{52}& \color[RGB]{40,160,70}\faTimes & \color[RGB]{200,85,80}\ding{52} \\
DISP \cite{zhou2019learning} & Token-level detection model & \color[RGB]{200,85,80}\ding{52} & \color[RGB]{200,85,80}\ding{52} & \color[RGB]{200,85,80}\ding{52} \\
FGWS \cite{mozes2021frequency}& Frequency-based word substitution & \color[RGB]{200,85,80}\ding{52} & \color[RGB]{200,85,80}\ding{52} & \color[RGB]{40,160,70}\faTimes \\
ADFAR \cite{bao2021defending}& Sentence-level detection model &\color[RGB]{200,85,80}\ding{52}  & \color[RGB]{200,85,80}\ding{52} & \color[RGB]{200,85,80}\ding{52} \\
RDE \cite{yoo2022detection}& Feature-based density estimation & \color[RGB]{200,85,80}\ding{52} & \color[RGB]{40,160,70}\faTimes & \color[RGB]{40,160,70}\faTimes \\ 
UAPAD (Ours) & Universal adversarial perturbation &  \color[RGB]{40,160,70}\faTimes&  \color[RGB]{40,160,70}\faTimes& \color[RGB]{40,160,70}\faTimes  \\ \hline
\end{tabular}}
\caption{Summary of previous detection methods in NLP system. Requiring clean/adv. data indicates what data is needed for the training and validation process. Requiring extra models indicates whether a separate new model needs to be trained for adversarial detection. Our approach is data-agnostic and can be easily integrated into the inference phase. }\label{tab:mothods}%
\vspace{-1.2em}
\end{table*}

In this work, we propose UAPAD, a novel framework to detect adversarial samples without exposure to training data and maintain a time consumption consistent with normal inference. We visualize our detection framework in Figure \ref{fig:1}.
Universal adversarial perturbations (UAPs) is an intriguing phenomenon on neural models, i.e. a single perturbation that is capable to fool a DNN for most natural samples \cite{zhang2021survey}, and can be calculated without the original training data \cite{mopuri2018generalizable, zhang2021towards}. 
We explore the utilization of UAPs to detect adversarial attacks, where adversarial and clean samples exhibit differential resistance to pre-trained perturbations on a sensitive feature subspace. 

Experimental results demonstrate that our training-data-agnostic method achieves promising detection accuracy with BERT on multiple adversarial detection tasks without using training or adversarial data, consuming additional inference time, or conducting overly extensive searches for hyper-parameters. Our main contributions are as follows:
\begin{itemize}
    \item We analyze and verify the association between adversarial samples and an intrinsic property of the model, namely UAPs, to provide a new perspective on the effects of adversarial samples on language models.
    \item We propose a novel framework (\textbf{UAPAD}), which efficiently discriminates adversarial samples without access to training data, and maintains an equivalent time consumption to normal inference. Our \textit{codes}\footnote{\url{https://github.com/SleepThroughDifficulties/UAPAD.git}} are publicly available.
\end{itemize}

\begin{figure}[t]
\centerline{\includegraphics[width=0.4\textwidth]{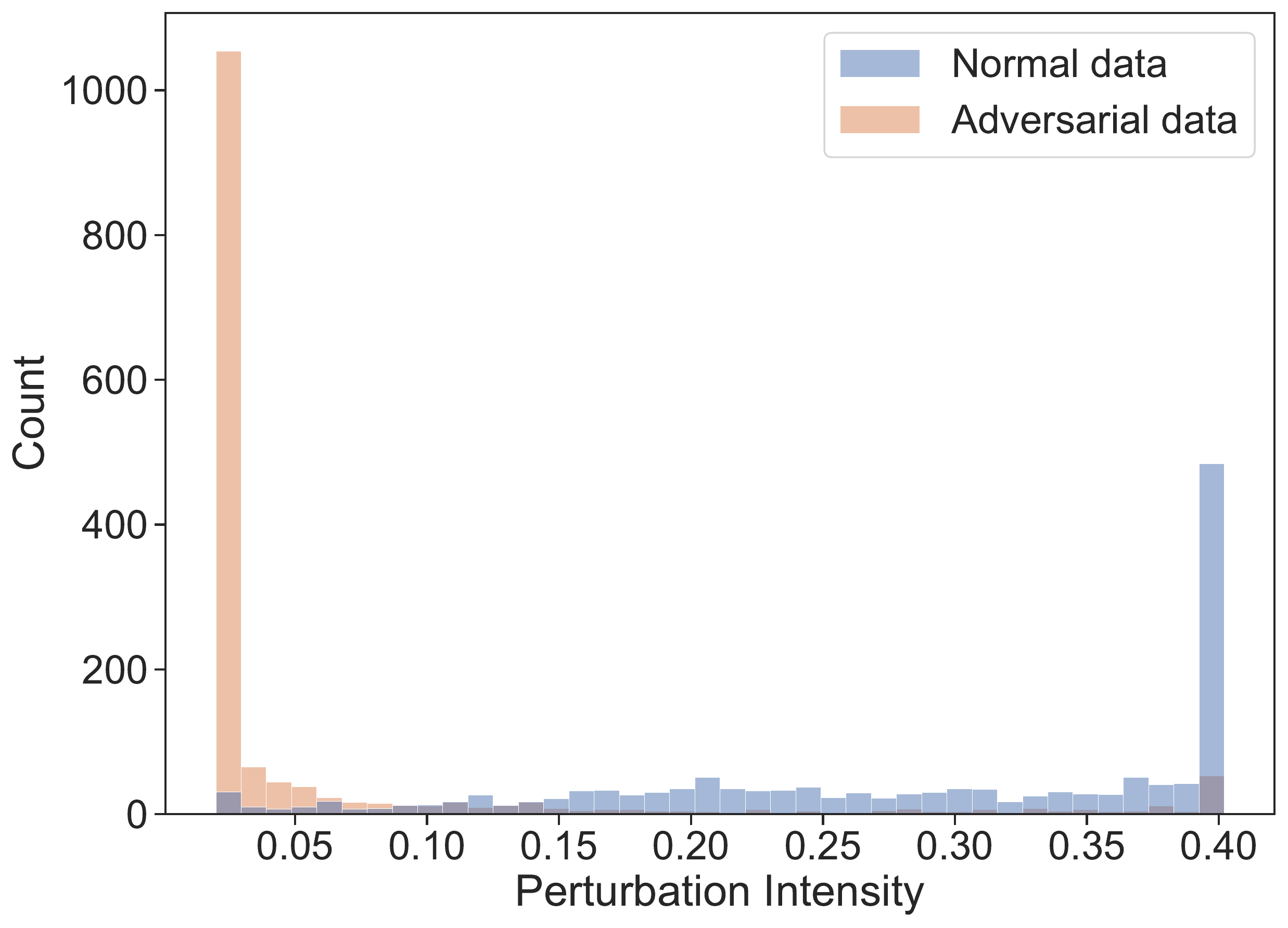}}
\caption{Illustration of the different resistance to universal perturbations in adv and clean data. Predictions for adversarial samples are inverted by a small perturbation intensity while clean samples maintain the original results.}
\label{fig:2}
\vspace{-1.2em}
\end{figure}
\section{Related Work}

\subsection{Universal adversarial perturbation}
The existence of UAPs has first been demonstrated by \cite{moosavi2017universal}, that a single perturbation can fool deep models when added to most natural samples. Such phenomena have been extensively verified in image \cite{khrulkov2018art}, text \cite{song2021universal}, and audio models \cite{li2020universal}. Some works attribute the existence of UAPs to a specific low-dimensional subspace, which is perpendicular to the decision boundary for most of the data. The attention on UAPs mainly focused on their construction, detection and defence \cite{zhang2021survey}, and neglected to explore the relationship between adversarial samples and UAPs. Our experimental results in Figure \ref{fig:2} demonstrate the tight connection between these two phenomena.

\subsection{Adversarial detection in NLP}
Adversarial detection is an emerging area of research on language model security.
A series of works analyze the frequency characteristics of word substitutions in pre-collected adversarial sentences and replace \cite{zhou2019learning, mozes2021frequency} or mask \cite{mosca2022suspicious} them to observe model reactions.
These methods rely on empirically designed word-level perturbations, which limit their generalizability across different attacks. \citet{ma2018characterizing} first proposed to train additional discriminative models to decide whether an input sentence has suffered from word-level adversarial substitution.
This idea was generalized by \citet{liu2022detecting} and \citet{yoo2022detection}, which determine the likelihood of a sentence has been perturbed.
However, they still require the statistical characteristics of the training data.
In this paper, we for the first time propose to construct data-agnostic models and achieve remarkable detection results.
\section{Method}
This section shows how to calculate the UAPs for a specific text model without obtaining training data.
And subsequently, how to detect adversarial data by pre-trained UAPs.

\paragraph{Data-free UAPs} We compute UAPs for a fine-tuned model by perturbing the substitute inputs, based on the fact that UAPs are generalized properties for a given model. 
We start with a parameter-frozen target network $f$ and a random perturbation $\delta$. 
The optimal situation is we can obtain some data that are involved in the training procedure. However, there are situations that we cannot access to training samples or it is unclear whether the accessible data is within the training set. To demonstrate the effectiveness of UAPAD under the data-agnostic scenario. We initialize the input embedding by randomly selecting data from an unrelated substitute dataset (e.g., the MNLI dataset in our experiments). 
It is a reasonable assumption that a defender can access a moderate amount of substitute data. 
These embeddings are subsequently updated to ensure the model's confidence score is above the threshold on them. In our framework, we only retain samples with model confidence above 85\% to calculate UAPs. 
We then optimize the perturbation $\delta$ by gradient-ascending the overall loss when added to all the inputs and project it to a normalized sphere of fixed radius to constrain its norm. 
We obtain a reasonable UAP when most predictions are induced to a fixed result under perturbation.

\paragraph{Adversarial Detection with UAPs} In Figure \ref{fig:2}, we illustrate the different resistance to UAPs between clean and adversarial samples. 
We utilize this property to conduct adversarial detection. 
Given an input $x$, we perform one inference on model $f$ to obtain the normal output $y = f(x)$ and perform another one when $x$ is perturbed by a calculated UAP $\delta$, that is $y' = f(x + w * \delta)$, where $w$ is a hyperparameter controlling the perturbation's intensity. 
We detect the input as an adversarial sample when $y \neq y'$. 
Noting that these two inferences can be computed in parallel, our approach does not introduce growth in inference time.
\section{Experimental Setup}

We experimentally validate our method on three well-accepted benchmarks: SST-2 \cite{socher2013recursive}, IMDB \cite{maas2011learning}, and AGNews \cite{zhang2015character}. The statistics of involved benchmark datasets are summarised in Appendix \ref{app: dataset}. We use the BERT-base \cite{devlin2018bert} as the target model and pre-generate adversarial samples for the detection task with three attack methods: TextFooler \cite{jin2020bert}, PWWS \cite{ren2019generating}, and BERTAttack \cite{li2020bert}. 

\subsection{Detecting Scenarios}\label{sec: 4.1}
Adversarial detection task requires a dataset $\mathcal{D}$, containing both clean samples $\mathcal{D}_{clean}$ and adversarial samples $\mathcal{D}_{adv}$. In the previous works, there exist two different strategies to construct adversarial datasets. Scenario $1$ (easy): The adversarial dataset consists of only successful attack samples. Scenario $2$ (hard): The adversarial dataset contains both successful and unsuccessful attack samples. Scenario 2 presents more challenging requirements for detection methods and is closer to real-world settings. We conduct experiments in both scenarios to fully illustrate the performance of UAPAD.

\subsection{Implementation Details}
We fine-tuned BERT using consistent settings with \cite{devlin2018bert}. 
For all three datasets, we took $1500$ training samples and saved their attack results under different attack algorithms as adversarial samples. 
UAPAD has a single hyperparameter $w$ (strength of universal perturbation), which we set to $0.5$ for all our detection experiments. 
Although we believe that a better weight exists and can boost the detection performance, we refuse to extend hyper-parameter searching which is against our original purpose. 
More implementation details and hyperparameters can be found in Appendix \ref{app: expsetup}. 

\paragraph{Evaluation Metrics} We use two metrics to measure our experimental results. \textbf{Detection accuracy (ACC)} measures the accuracy of classification results on all samples, and \textbf{F1-score (F1)} measures the harmonic mean of precision and recall scores. 
Similar to DISP, our method provides a direct discriminant rather than a score and therefore does not apply to the AUC metric. 

\paragraph{Baselines} We compare our proposed methods with four strong baselines. Details are summarized in Appendix \ref{app: baselines}.
\begin{itemize}
    \item \textbf{MLE} \cite{lee2018simple} proposes to train detection models based on Mahalanobis distance.
    \item \textbf{DISP} \cite{zhou2019learning} verifies the likelihood that a token has been perturbed.
    \item \textbf{FGWS} \cite{mozes2021frequency} substitutes low-frequency words in the sentence to detect Word-level attacks.
    \item \textbf{RDE} \cite{yoo2022detection} models the probability density of inputs and generates the likelihood of a sentence being perturbed.
\end{itemize}


\section{Experiment Results and Discussions}
In this section, we show the experimental performance of our proposed method under the two scenarios in Section \ref{sec: 4.1}, and investigate different defence methods on the inference time consumption.

\begin{table}[t]
\centering
\resizebox{0.97\columnwidth}{!}{%
\begin{tabular}{l|l|cc|cc|cc}
\hline\hline
 &  & \multicolumn{2}{c|}{TextFooler} & \multicolumn{2}{c|}{PWWS} & \multicolumn{2}{c}{BERT-ATTACK} \\ \cline{3-8} 
\multirow{-2}{*}{Datasets} & \multirow{-2}{*}{Methods} & Acc & F1 & Acc & F1 & Acc & F1 \\ \hline
& MLE & 79.6& 77.0& 77.5& 77.2& 70.3& 63.7 \\
 & DISP & 71.2 & 66.0 & 74.4 & 70.9 & 70.8 & 65.4 \\
 & FGWS & 70.2 & 63.5 & 82.5 & \textbf{81.3} & 70.3 & 63.7 \\
 & RDE & 78.0 & 73.4 & 79.5 & 77.6 & 83.4 & 81.3 \\
\multirow{-4}{*}{SST-2} & \cellcolor[HTML]{D1F39D}UAP (Ours) & \cellcolor[HTML]{D1F39D}\textbf{83.7} & \cellcolor[HTML]{D1F39D}\textbf{83.1} & \cellcolor[HTML]{D1F39D}\textbf{82.5} & \cellcolor[HTML]{D1F39D}80.9 & \cellcolor[HTML]{D1F39D}\textbf{87.4} & \cellcolor[HTML]{D1F39D}\textbf{87.6} \\ \hline
& MLE & 83.7&\textbf{ 81.8}& \textbf{81.5}& \textbf{79.4}& 83.7& \textbf{82.3} \\
 & DISP & 68.8 & 70.6 & 66.8 & 68.2 & 67.3 & 68.8 \\
 & FGWS & 74.7 & 69.7 & 77.5 & 74.0 & 74.4 & 69.3 \\
 & RDE & 83.2 & 75.6 & 82.0 & 74.4 & 83.5 & 76.6 \\
\multirow{-4}{*}{IMDB} & \cellcolor[HTML]{D1F39D}UAP (Ours) & \cellcolor[HTML]{D1F39D}\textbf{84.1} & \cellcolor[HTML]{D1F39D}72.6 & \cellcolor[HTML]{D1F39D}81.2 & \cellcolor[HTML]{D1F39D}73.3 & \cellcolor[HTML]{D1F39D}\textbf{83.8} & \cellcolor[HTML]{D1F39D}78.4 \\ \hline
& MLE & 79.9& 79.6& 77.3& 76.9& 82.7& 78.6 \\
 & DISP & 86.7 & 86.4 & 86.9 & 86.6 & 83.5 & 82.6 \\
 & FGWS & 68.3 & 59.6 & 75.0 & 70.6 & 68.2 & 59.4 \\
 & RDE & 85.0 & 86.7 & 85.8 & 81.4 & 88.2 & 88.0 \\
\multirow{-4}{*}{AGNEWS} & \cellcolor[HTML]{D1F39D}UAP (Ours) & \cellcolor[HTML]{D1F39D}\textbf{95.8} & \cellcolor[HTML]{D1F39D}\textbf{95.6} & \cellcolor[HTML]{D1F39D}\textbf{94.4} & \cellcolor[HTML]{D1F39D}\textbf{93.1} & \cellcolor[HTML]{D1F39D}\textbf{94.9} & \cellcolor[HTML]{D1F39D}\textbf{94.9} \\ \hline\hline
\end{tabular}}
\caption{Adversarial detection results on easy scenario.}
\label{tab:2}
\vspace{-0.7em}
\end{table}

\subsection{Main Results}
Table \ref{tab:2} and \ref{tab:3} show the detect results on three datasets and three attacks. 
The highest means are marked in \textbf{bold}. 
Out of the 18 combinations of dataset-attack-scenario, UAPAD achieves the best performance on 15 of them on ACC and 12 of them on F1 metric, 
which demonstrates the competitiveness of our data-agnostic approach. 
UAPAD guarantees remarkable detection performance on the SST-2 and AGNews datasets and suffers from a small degradation on the IMDB dataset. 
We argue that the average length of sentences is greater on IMDB, resulting in stronger dissimilarity between the adversarial sample generation by attack algorithms and the original sentence. 
On the AGNews dataset, UAPAD provided a 3-11\% increase in detection accuracy relative to the baseline approach. 
We attribute this impressive improvement to more categories on this task, which improved the accuracy of estimation on the model's UAPs.

\begin{table}[t]
\centering
\resizebox{0.97\columnwidth}{!}{%
\begin{tabular}{l|l|cc|cc|cc}
\hline\hline
 &  & \multicolumn{2}{c|}{TextFooler} & \multicolumn{2}{c|}{PWWS} & \multicolumn{2}{c}{BERT-ATTACK} \\ \cline{3-8} 
\multirow{-2}{*}{Datasets} & \multirow{-2}{*}{Methods} & Acc & F1 & Acc & F1 & Acc & F1 \\ \hline
& MLE & 76.7& 75.2& 77.2& 77.9& 82.4& 83.0 \\
 & DISP & 71.2 & 64.9 & 74.4 &68.8 & 70.7 & 64.9 \\
 & FGWS & 65.7 & 55.6 & 69.3 & 61.7 & 64.6 & 53.5 \\
 & RDE & 77.4& 73.2& 78.6& 77.7& 82.9& 81.0 \\
\multirow{-4}{*}{SST-2} & \cellcolor[HTML]{D1F39D}UAP (Ours) & \cellcolor[HTML]{D1F39D}\textbf{80.7} & \cellcolor[HTML]{D1F39D}\textbf{80.9} & \cellcolor[HTML]{D1F39D}\textbf{78.8} & \cellcolor[HTML]{D1F39D}\textbf{78.1} & \cellcolor[HTML]{D1F39D}\textbf{84.9} & \cellcolor[HTML]{D1F39D}\textbf{85.1} \\ \hline
& MLE & 74.0 & \textbf{82.4} &74.5&76.2& 74.4& 78.9 \\
 & DISP & 64.1 & 53.7 &62.4 &51.9 &63.2 &52.6  \\
 & FGWS &62.1 &47.4&63.5&49.8&58.6&39.6 \\
 & RDE & 77.4&73.2&\textbf{78.6}&77.7& \textbf{82.9}& \textbf{81.3}\\
\multirow{-4}{*}{IMDB} & \cellcolor[HTML]{D1F39D}UAP (Ours) & \cellcolor[HTML]{D1F39D}\textbf{78.3} & \cellcolor[HTML]{D1F39D}76.8 & \cellcolor[HTML]{D1F39D}77.5 & \cellcolor[HTML]{D1F39D}\textbf{76.8} & \cellcolor[HTML]{D1F39D}79.3 & \cellcolor[HTML]{D1F39D}78.0 \\ \hline
& MLE & 77.9 & 77.0 & 73.2 & 71.5 & 79.2 & 78.9 \\
 & DISP & 86.1 & 84.5 & 85.4 & 81.0 & 83.1 & 81.5 \\
 & FGWS &64.7 & 52.8 & 67.7 & 58.4 & 64.1 & 51.6 \\
 & RDE & 85.1 & 84.4 & 77.0 & 78.5 & 86.6 & 85.7 \\
\multirow{-4}{*}{AGNEWS} & \cellcolor[HTML]{D1F39D}UAP (Ours) & \cellcolor[HTML]{D1F39D}\textbf{88.6} & \cellcolor[HTML]{D1F39D}\textbf{88.1} & \cellcolor[HTML]{D1F39D}\textbf{86.3 }& \cellcolor[HTML]{D1F39D}\textbf{81.4} & \cellcolor[HTML]{D1F39D}\textbf{92.0} & \cellcolor[HTML]{D1F39D}\textbf{91.9} \\ \hline\hline
\end{tabular}}
\caption{Adversarial detection results on hard scenario.}
\label{tab:3}
\vspace{-0.7em}
\end{table}

\subsection{Time Consumption}
To further reveal the strength of UAPAD besides its detection performance, we compare its GPU training time consumption with other baseline methods.
As is demonstrated in Table \ref{tab:4}, The time consumption of UAPAD is superior to all the comparison methods. 
Only FGWS \cite{mozes2021frequency} exhibits similar efficiency to ours (with about 20\% time growth on SST-2 and IMDB). 
FGWS neither contains a backpropagation process in the inference phase, but still requires searching the pre-built word list for substitution.

\begin{table}[h]
\centering
\resizebox{0.7\columnwidth}{!}{%
\begin{tabular}{l|c|c}
\hline \hline
\textbf{Methods} & \textbf{SST-2} & IMDB \\ \hline
finetune & 3.42 & 4.33 \\ \hline
\textbf{UAP (Ours)} & \textbf{3.58} & \textbf{4.61} \\ \hline
DISP & 19.2 & 24.6 \\ \hline
FGWS & 4.17 & 5.58 \\ \hline
RDE & 76.8 & 102.3 \\ \hline\hline
\end{tabular}}
\caption{GPU time consumption (seconds) of detection 1500 samples. UAPAD costs nearly the same as normal predictions.}
\label{tab:4}
\vspace{-1em}
\end{table}

\section{Conclusion}
In this paper, we propose that adversarial samples and clean samples exhibit different resistance to UAPs, a model-related vector that can be calculated without accessing any training data. 
Based on this discovery, we propose UAPAD as an efficient and application-friendly algorithm to overcome the drawbacks of previous adversarial detection methods in terms of slow inference and the requirement of training samples. UAPAD acts by observing the feedback of inputs when perturbed by pre-computed UAPs. 
Our approach achieves impressive detection performance against different textual adversarial attacks in various NLP tasks. 
We call for further exploration of the connection between adversarial samples and UAPs.

\section*{Acknowledgements}
The authors wish to thank the anonymous reviewers for their helpful comments. This work was partially funded by National Natural Science Foundation of China (No.61976056,62076069) and Natural Science Foundation of Shanghai (23ZR1403500).

\section{Limitations}
This section discusses the potential limitations of our work. 
This paper's analysis of model effects mainly focuses on common benchmarks for adversarial detection, which may introduce confounding factors that affect the stability of our framework. 
Our model's performance on more tasks and more attack algorithms is worth further exploring. 
Our detection framework exploits the special properties exhibited by the adversarial sample under universal perturbation. 
We expect a more profound exploration of improving the connection between UAPs and adversarial samples. 
In Figure \ref{fig:2}, we note that a small number (about $3\%$) of clean and adversarial samples do not suffer from UAP interference. 
It is worth conducting an analysis of them to further explore the robustness properties of the language models.
We leave these problems to further work.

\bibliography{anthology,custom}
\bibliographystyle{acl_natbib}

\newpage
\appendix
\section{Dataset Statistics}
\label{app: dataset}
\begin{table}[h]
\begin{tabular}{@{}cccc@{}}
\toprule
\multicolumn{1}{l}{\textbf{Dataset}} & \multicolumn{1}{l}{\textbf{Train/Test}} & \multicolumn{1}{l}{\textbf{Classes}} & \multicolumn{1}{l}{\textbf{\#Words}} \\ \midrule
\textbf{SST-2}                       & 67k/1.8k                                & 2                                    & 19                                   \\
\textbf{IMDB}                        & 25k/25k                                 & 2                                    & 268                                  \\
\textbf{AGNews}                     & 120k/7.6k                               & 4                                    & 40                                   \\ \bottomrule
\end{tabular}
\caption{Statistics of datasets. In our experiments, we partition an additional 10 percent of the training set as the validation set to calculate the DSRM of the model.}
\end{table}

\section{Experimental Details}
\label{app: expsetup}
In this appendix, we show the hyper-parameters used for our proposed method.
We fine-tune the BERT-base model by the official default settings. 
For SST-2, we use the official validation set, while for IMDB and AGNews, we use $10\%$ of the data in the training set as the validation set.
The validation set and the adversarial samples generated using the validation set are used to select hyper-parameters.
All three attacks are implemented using TextAttack\footnote{https://github.com/QData/TextAttack} with the default parameter settings.
Following \citet{zhou2019learning}, for SST-2, IMDB and AGNews, we build a balanced set consisting of $1500$ clean test samples and $1500$ adversarial samples to evaluate our proposed methods and all the baselines in this paper.
We train our models on NVIDIA RTX 3090 GPUs (four for RDE and one for other methods). All experiments are run on three different seeds and report the mean result.

\section{Baseline Details}
\label{app: baselines}
We compare our proposed detectors with three strong baselines in adversarial example detection.

\textbf{MLE} \citep{lee2018simple}:
A simple yet effective method for detecting OOD and adversarial examples in the image processing domain. The main idea is to induce a generative classifier under Gaussian discriminant analysis, resulting in a detection score based on Mahalanobis distance.

\textbf{DISP} \citep{zhou2019learning}: A novel BERT-based framework can identify perturbations and correct malicious perturbations. it contains two independent components, a perturbation discriminator and an estimator for token recovery. To detect adversarial attacks, the discriminator verifies the likelihood that a token in the sample has been perturbed.

\textbf{FGWS} \cite{mozes2021frequency} leverages the frequency properties of adversarial word substitution for the detection of adversarial samples. Briefly, FGWS replaces the low-frequency words with their most frequent synonyms in the dictionary to detect the perturbation.

\textbf{RDE} \cite{yoo2022detection} proposes a competitive adversarial detector based on density estimation. RDE models the probability density of the entire text and generates the likelihood of a text being perturbed.

\end{document}